\documentclass[letterpaper, 10 pt, conference]{ieeeconf}  % Comment this line out if you need a4paper

\IEEEoverridecommandlockouts                              % This command is only needed if 
\usepackage{graphics} % for pdf, bitmapped graphics files
\usepackage{epsfig} % for postscript graphics files
\usepackage{subfigure} % for postscript graphics files
\usepackage{amsmath} % assumes amsmath package installed
\usepackage{bm}
\usepackage{amssymb}  % assumes amsmath package installed
\usepackage{cite}

\let\labelindent\relax
\usepackage{amsfonts}
\usepackage{cuted}
\usepackage{doi}
\usepackage[inline]{enumitem}
\usepackage{adjustbox}
\usepackage{array}
\usepackage{hyperref}

\title{\LARGE \bf
HumanMimic: Learning Natural Locomotion and Transitions for Humanoid Robot via Wasserstein Adversarial Imitation
}

\author{Annan Tang$^{1}$, Takuma Hiraoka$^{1}$, Naoki Hiraoka$^{1}$, Fan Shi$^{1,2}$, Kento Kawaharazuka$^{1}$, \\ Kunio Kojima$^{1}$, Kei Okada$^{1}$ and Masayuki Inaba$^{1}$% <-this % stops a space
\thanks{$^{1}$JSK Lab, Graduate School of Information Science and Technology,
The University of Tokyo, 7-3-1 Hongo, Bunkyo-ku, Tokyo 113-8656, Japan. 
        {\tt\small tang@jsk.imi.i.u-tokyo.ac.jp}}%
\thanks{$^{2}$AI Center, ETH Zürich, 8092 Zürich, Switzerland.
        {\tt\small fan.shi@ai.ethz.ch}}%
}

\begin{document}

\maketitle
\thispagestyle{empty}
\pagestyle{empty}

\begin{abstract}
Transferring human motion skills to humanoid robots remains a significant challenge. In this study, we introduce a Wasserstein adversarial imitation learning system, allowing humanoid robots to replicate natural whole-body locomotion patterns and execute seamless transitions by mimicking human motions. First, we present a unified primitive-skeleton motion retargeting to mitigate morphological differences between arbitrary human demonstrators and humanoid robots. An adversarial critic component is integrated with Reinforcement Learning (RL) to guide the control policy to produce behaviors aligned with the data distribution of mixed reference motions. Additionally, we employ a specific Integral Probabilistic Metric (IPM), namely the Wasserstein-1 distance with a novel soft boundary constraint to stabilize the training process and prevent mode collapse. Our system is evaluated on a full-sized humanoid JAXON in the simulator. The resulting control policy demonstrates a wide range of locomotion patterns, including standing, push-recovery, squat walking, human-like straight-leg walking, and dynamic running. Notably, even in the absence of transition motions in the demonstration dataset, the robot showcases an emerging ability to transit naturally between distinct locomotion patterns as desired speed changes. Supplementary video can be found here: \href{https://www.youtube.com/watch?v=sdM11yHpzi8}{WATCH VIDEO}.
\end{abstract}
%%%%%%%%%%%%%%%%%%%%%%%%%%%%%%%%%%%%%%%%%%%%%%%%%%%%%%%%%%%%%%%%%%%%%%%%%%%%%%%%%%%%%%%
\section{Introduction}
Natural selection has shaped human ability, enabling humans to perform various locomotion behaviors and adeptly shift gait patterns in response to speed changes or external disturbances. Transferring the natural-looking locomotion and seamless transitions to humanoid robots remains a longstanding challenge, primarily due to the control complexity and intricacies of motion design. 

While numerous studies based on simplified models\cite{miura2011human}, \cite{wensing2013high}, \cite{kamioka2017dynamic}, \cite{sugihara20213d} and optimal control\cite{ishihara2019full}, \cite{chignoli2021humanoid} have demonstrated promising performance on structured locomotion paradigms, the intrinsically under-actuated and nonlinear characteristics of humanoids complicate the establishment of a unified model that accurately captures the dynamics across diverse gait transitions. On the other hand, deep reinforcement learning (deep RL) semi-automates the complex modeling process by maximizing the cumulative reward, leading to its growing popularity in developing advanced locomotion skills for quadrupedal robots\cite{hwangbo2019learning}, \cite{hoeller2023anymal}, bipedal robots\cite{xie2020learning}, \cite{siekmann2021sim} and even humanoid robots\cite{radosavovic2023learning}, \cite{kim2023torque}. Nevertheless, RL-generated motions for high-DOF humanoids often exhibit undesired whole-body behaviors, including irregular arm swings, aggressive ground impacts, and unnatural gaits. Typical solutions utilize supplemental footstep planners\cite{singh2022learning}, heuristic gait reward design\cite{siekmann2021sim} or pre-optimized gait and joint trajectory\cite{crowley2023optimizing} to induce specific locomotion patterns. But given our limited understanding of the underlying characteristics that depict the natural behaviors of humans, these modules frequently employ basic principles like symmetry and energy minimization\cite{yu2018learning}, resulting in more stereotypical robotic motions compared to humans.
% While such behaviors are dynamics-feasible in simulations, they often prove impractical in real-world scenarios.
\begin{figure}[t]
        \centering
        \includegraphics[width=0.42\textwidth]{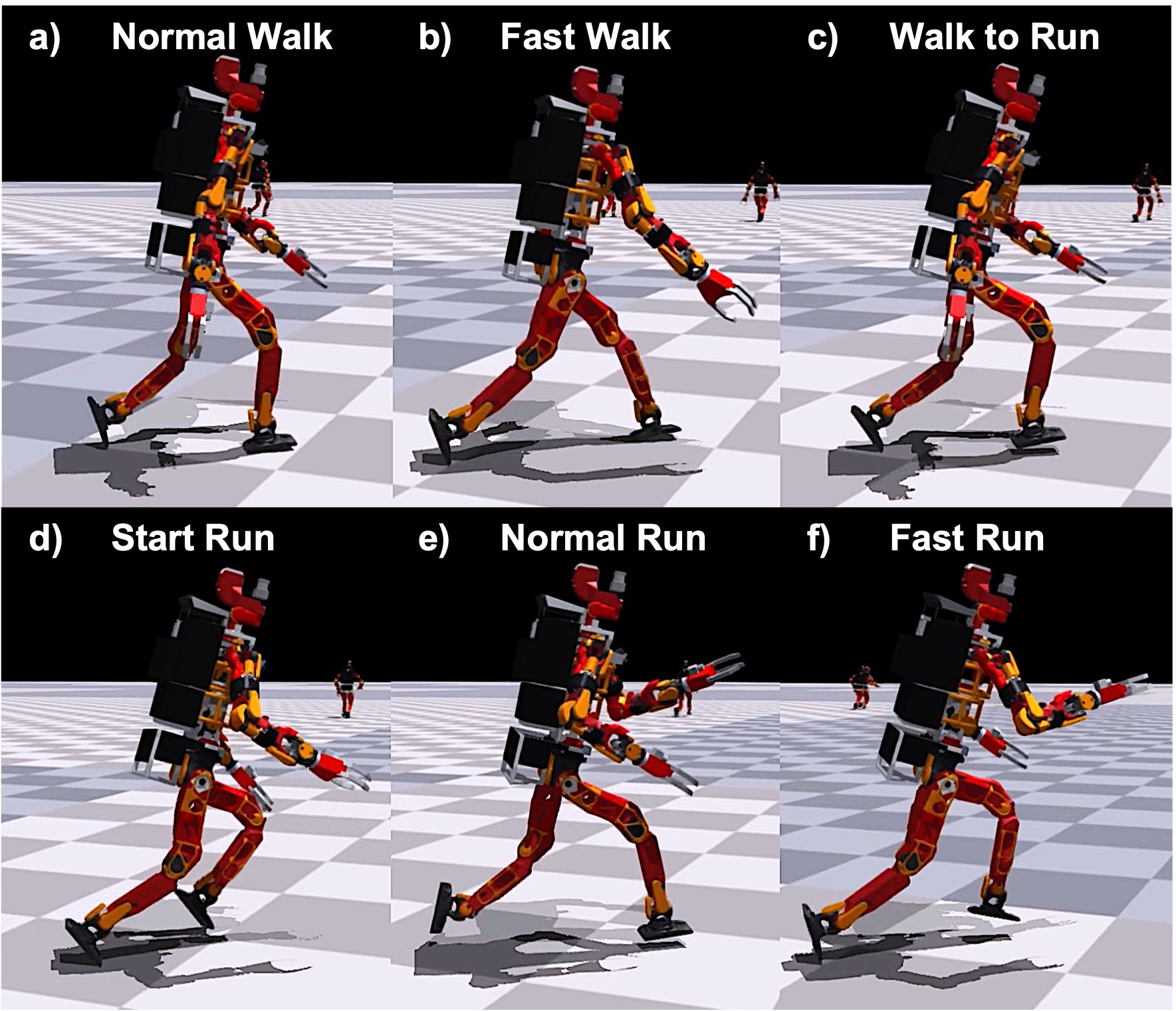} 
        \caption{Our Wasserstein adversarial imitation learning system enables a full-sized humanoid to exhibit various human-like natural locomotion behaviors and achieve seamless transitions as velocity command changes.}
        \label{fig:intro}
\end{figure}

% Recently, adversarial motion prior (AMP)\cite{} offers an appealing solution to acquire natural motions without laborious reward engineering. AMP leverages an additional discriminator that outputs a style reward to encourage generated motions to resemble human demonstrations. In practice, the discriminator trained with binary cross entropy (BCE) or Least-Square (LS) loss commonly suffers from unstable training and mode collapse due to the improper metric to measure the distance between two non-overlapping probability distributions in high dimensional space. Furthermore, the significant morphological discrepancy between humans and robots complicates the direct imitation of the Motion capture (Mocap) data \cite{}.

For acquiring natural motions without the need for laborious reward engineering, the adversarial motion prior (AMP)\cite{peng2020learning} exploits an additional discriminator that outputs a style reward to encourage generated motions to resemble human demonstrations. In practice, discriminators trained with binary cross entropy (BCE) or least-square (LS) loss often face unstable training and mode collapse, mainly due to the inadequacy of the metrics used to measure distances between non-overlapping probability distributions in high-dimensional spaces. In the closely related domain of generative adversarial networks (GANs), researchers have introduced several types of integral probability metrics (IPMs)\cite{sriperumbudur2009integral}, \cite{li2017mmd}, \cite{mroueh2017fisher}, especially the Wasserstein distance\cite{gulrajani2017improved}, \cite{dadashi2021primal}, \cite{durugkar2021adversarial}, to address the challenges above. However, the unbounded Wasserstein distance \cite{li2023learning} presents a significant challenge when trying to balance the style reward with other task-specific rewards like velocity tracking. Moreover, the significant morphological differences between human demonstrators and humanoid robots, including joint configurations, body proportions, and bone hierarchies, pose challenges for the direct imitation of human demonstrations.

In this work, we present an adversarial imitation learning system that enables full-sized humanoids to autonomously acquire a variety of realistic locomotion behaviors through imitating human demonstrations. First, we introduce a unified primitive-skeleton motion retargeting approach to address morphological differences between arbitrary human demonstrators and humanoid robots. We exploit the power of the Wasserstein-1(W-1) distance, incorporating a novel soft boundary constraint, to ensure stable training dynamics and prevent the convergence of generated motions to a limited set of trivial modes. The learned one policy showcases a diverse array of robust and natural locomotion patterns, encompassing standing, push-recovery, squat walking, human-like straight-leg walking, dynamic running, and seamless transitions in response to changes in velocity commands, as shown in Figure \ref{fig:intro}. In short, our main contributions are: 
\begin{enumerate*}[label=(\roman*)]
        \item Proposing an improved adversarial imitating learning system with
        Wasserstein critic and soft boundary constraints to address unstable training and mode collapse.
        \item Detailing a unified primitive-skeleton motion retargeting technique applicable to arbitrary human skeleton sources and humanoid models.
        \item Achieving the whole-body natural locomotion and transitions for humanoids and evaluating the robustness through sim-to-sim settings in a high-fidelity simulator.
\end{enumerate*}

\section{Related Works}

\textbf{RL for bipedal locomotion}. Recent advancements in RL-based control strategies have significantly enhanced bipedal locomotion\cite{lirobust}, \cite{siekmann2021blind}. For instance, the bipedal robot Cassie not only mastered versatile gait patterns through the use of periodic-parameterized reward functions\cite{siekmann2021sim} but also achieved the Guinness World Record for the fastest 100m dash using pre-optimized reference running gaits\cite{crowley2023optimizing}. Jeon et al.\cite{jeon2023benchmarking} utilized potential-based reward shaping to ensure faster convergence and more robust humanoid locomotion. Shi et al.\cite{shi2022reference} integrated an assistive force curriculum into the learning process, allowing the acquisition of multiple agile humanoid motion skills in reference-free settings. In a more recent study, the full-sized humanoid HRP-5P\cite{singh2023learning} showcased robust walking using actuator current feedback, while Kim et al.\cite{kim2023torque} demonstrated a torque-based policy to bridge sim-to-real gaps. DeepMind\cite{haarnoja2023learning} managed to instill agile soccer skills in a miniature humanoid via a two-stage teacher-student distillation and self-play. Additionally, attention-based transformers \cite{radosavovic2023learning} have been employed to achieve more versatile locomotion in the humanoid Digit.

\textbf{Motion imitation from real-world demonstrations}. Leveraging reference motions from creatures enables robots to acquire natural and versatile locomotion skills\cite{bohez2022imitate},  \cite{han2023lifelike} that are challenging to define manually. Predominant imitation strategies involve tracking either reference joint trajectories\cite{peng2018deepmimic}, \cite{peng2020learning}, \cite{crowley2023optimizing} or extracted gait features\cite{yin2021run}, \cite{kang2022animal}. However, these explicit tracking techniques, often limited to separate motion clips, can disrupt smooth transitions between different locomotion patterns. Drawing inspiration from generative adversarial imitation learning (GAIL) \cite{ho2016generative}, Peng et al. introduced AMP\cite{peng2021amp} and successor ASE \cite{peng2022ase}. These approaches empower physics-based avatars to carry out objective tasks while simultaneously imitating the underlying motion styles from extensive unstructured datasets in an implicit manner. Variants of AMP have been further employed for learning agile quadrupedal locomotion\cite{li2023learning}, \cite{escontrela2022adversarial}, \cite{vollenweider2023advanced} and terrain-adaptive skills \cite{wu2023learning}, \cite{wang2023amp}, exemplifying its efficacy in eliminating the need for intricate reward function design. 

Despite the advancements in other domains, methods similar to AMP have not been extensively explored for humanoid robots. To bridge this gap, in this work, we present a Wasserstein adversarial imitation system with soft boundary constraints as an enhancement to the existing AMP techniques. We aim to provide a foundational training algorithm for future deployment on full-sized humanoid robots in real-world scenarios.
\begin{figure}[t]
        \centering
        \includegraphics[width=0.47\textwidth]{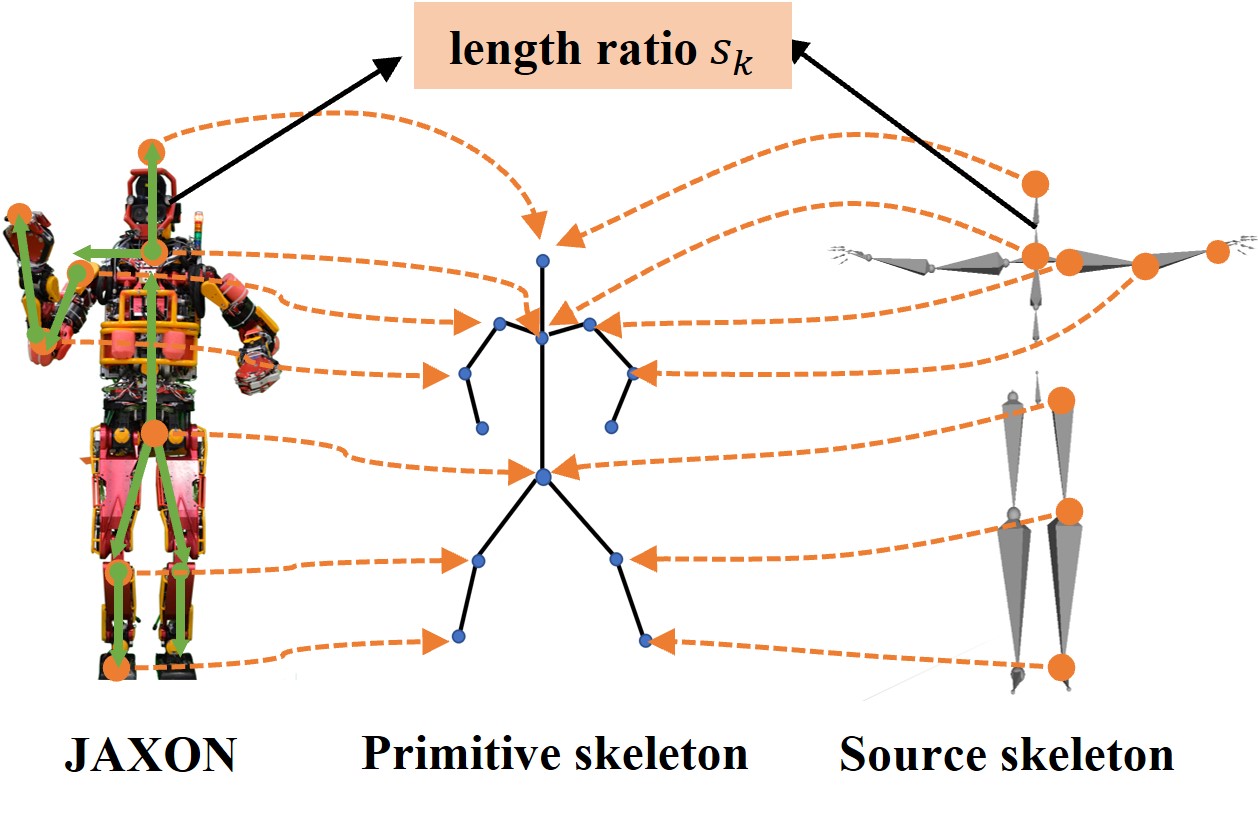} 
        \caption{Binding the humanoid JAXON and the MoCap skeleton involves merging their bones into a common primitive skeleton.}
        \label{fig:primitive_binding}
\end{figure}
\section{Motion Retargeting}
To transfer reference motion to the robot, certain retargeting methods\cite{ayusawa2017motion}, \cite{grandia2023doc} consider both kinematic and dynamic constraints, requiring accurate dynamic modeling or complex balance controllers. In this section, we detail a flexible motion retargeting approach based on the unified primitive skeleton, emphasizing geometry consistency. The kinematic and dynamic constraints such as feet contact state and balance will be satisfied in the reinforcement learning paradigm in the next section. Our retargeting involves four key procedures.

\textbf{Unified primitive skeleton binding}. Skeletal structures of humans and humanoid robots are known to correspond to homeomorphic graphs \cite{aberman2020skeleton}. Leveraging this property, we extract what we term \textit{primitive skeleton} that encapsulates the foundational geometric and hierarchical characteristics shared across various skeletons. In the process of primitive skeleton binding, we first construct kinematic trees for all involved skeletons. These trees are subsequently merged into a unified primitive skeleton, retaining only a single bone between two successive key joints. Users manually select $n$ key joints, offering an intuitive and flexible mechanism for loose binding between the source and target skeleton groups. Once binding is complete, we compute the length ratio $S = \{ s_k \mid k \in \{1, \ldots, n\} \}$ for each bone within the primitive skeleton. An illustrative example of this binding between the Humanoid JAXON \cite{kojima2015development} and CMU MoCap\cite{cmudata} skeleton is presented in Figure \ref{fig:primitive_binding}.

% \textbf{Coordinate transformation}. Each MoCap motion sequence $M_s = \{ m_t \mid t \in \{1, \ldots, T\} \}$ includes consecutive $ T $ motion frames. Each motion frame $ m_t = (^wP_r, ^wR_r, ^{0}R_1, \ldots, ^{j-1}R_j)$ consists of the root position $^wP_r$ and root orientation $^wR_r$ w.r.t. the world coordinate, as well as the local orientation $^{j-1}R_j$ of each source skeleton's joint $i$ w.r.t. its parent coordinate. Iterative homogeneous transformations are performed from the root towards the limbs and head to obtain the global position $^wP_j$ of each joint $j$ in source skeletons. The transformed motion frame can be represented as $ m^{\prime}_{t} = (^wP_r, ^wR_r, ^{w}P_1, \ldots, ^{w}P_j)$. Then we can easily calculate the relative position vector $\vec{r_k}$ of adjacent key joints $k$ and $k-1$ in source skeletons. We scale up the relative position vector $\vec{r_i}$ by ratio $s_k$ to acquire the relative position vector $\Vec{r_i}^{\prime}$ in target robot skeletons. Finally, we sum up the relative position vector $\Vec{r_i}^{\prime}$ along the kinematic chains and perform a coordinate transformation to get the positions of key joint $P_{k}$, end-effectors Cartesian position $P_{E}$ and quaternion $Q_{E}$ w.r.t the root of robot skeletons.
\textbf{Coordinate transformation}. We consider the MoCap dataset represented by the predefined skeletal kinematic tree $L^{\prime} = \{l^{\prime}_{ij}\}$ and the corresponding motion sequence of $T$ frames $M^{\prime}_s = \{ m^{\prime}_t \mid t \in \{1, \ldots, T\} \}$, with frame $t$ as $ m^{\prime}_t = (^wP_r^{\prime}, ^wR_r^{\prime}, ^{0}R_1^{\prime}, \ldots, ^{j-1}R_j^{\prime})$. Here, $l^{\prime}_{ij}$ is the bone length between directly connected joints $i$ and $j$, $^wP_r^{\prime}$ and  $^wR_r^{\prime}$ represent the root's position and orientation w.r.t the world coordinate, and $^{j-1}R_j^{\prime}$ indicates the local orientation of the source skeleton's joint $j$ w.r.t its parent joint. Applying iterative homogeneous transformations along the given kinematic tree $L^{\prime}$, denoted by ${ }^w P_j^{\prime}=H\left(^wP_r^{\prime}, ^wR_r^{\prime}, ^{0}R_1^{\prime}, \ldots, ^{j-1}R_j^{\prime}, L^{\prime}\right)$, we derive the global position for each joint in the source skeletons. The relative position vector between adjacent key joints is computed as $\vec{r_k}^{\prime}={ }^w P_k^{\prime}-{ }^w P_{k-1}^{\prime}$. We scale it by $\vec{r}_k=s_k \cdot \vec{r_k}^{\prime}$ to get the relative position vector $\vec{r}_k$ in target robot skeleton. Finally, we sum up the relative position vector along the kinematic chains and apply a transformation to get the key joint Cartesian positions w.r.t the root of robot skeletons as $^rP_k=H\left(\sum_{i=1}^k \vec{r}_i\right)$. All the end-effector poses $^rP_e \in \mathbb{R}^3 \times \mathbb{SO}(3)$ are incorporated into the final robot motion frame $m_t=\left({ }^w P_r,{ }^w R_r, ^rP_k, ^rP_e\right)$, here $e$ denotes wrists, feet and head.

\textbf{Multi-Objective inverse kinematics}. To map the key joint Cartesian position  $^rP_k$ and end-effector pose $^rP_e$  to joint positions $\theta = \left(\theta_1, \theta_2, \ldots, \theta_n\right)$, we formulate the whole-body inverse kinematics as a gradient-based optimization problem \cite{starke2018memetic} with three goals, 
\begin{equation}
        \begin{aligned}
        &C_1=\sum_k \left\|^rP_k-p_k\left(\theta\right)\right\|^2,\\
        &C_2=\sum_e \left\|^rP_e-p_e\left(\theta\right)\right\|^2,\\
        &C_3=\left\|\theta_t-\theta_{t-1}\right\|^2,\\
        \end{aligned}
\end{equation}
where the $p_k\left(\theta\right)$ and $p_e\left(\theta\right)$ are the calculated Cartesian position and pose for joints and end-effectors during gradient descent iterations. The main goals consist of the \textbf{position goal} $C_1$ for all key joints and the \textbf{pose goal} $C_2$ for the end-effectors including hands, foot soles and head. An additional \textbf{minimal displacement goal} $C_3$ is introduced to maintain each joint variable close to the previous motion frames. This is crucial for the highly redundant humanoids as multiple solutions might satisfy $C_1$ and $C_2$. The overall objective function is the weighted sum of each individual goal cost,
\begin{equation}
C = \arg min_\theta \sum_i \kappa_iC_i\left(\theta\right).
\end{equation}
The weights $\kappa_i$ are determined as ${(1, 1, 0.2)}$ heuristically. The joint position and velocity limitations are incorporated into the constraints,
\begin{equation}
        \begin{aligned}
        \theta_{\min } &\leq \theta_t \leq \theta_{\max }, \\
        \dot{\theta}_{\min } &\leq \frac{\theta_t - \theta_{t-1}}{\Delta t} \leq \dot{\theta}_{\max }.
        \end{aligned}
\end{equation}
        
\textbf{Post-Processing}. We compute root and joint velocities from sequential frame differences. Linear and Slerp interpolation are applied separately for position and orientation between discrete motion frames. Moreover, an exponential moving average filter is applied to smooth position and velocity spikes. 

\section{Wasserstein Adversarial Imitation}
\begin{figure*}[t]
        \centering
        \includegraphics[width=0.85\textwidth]{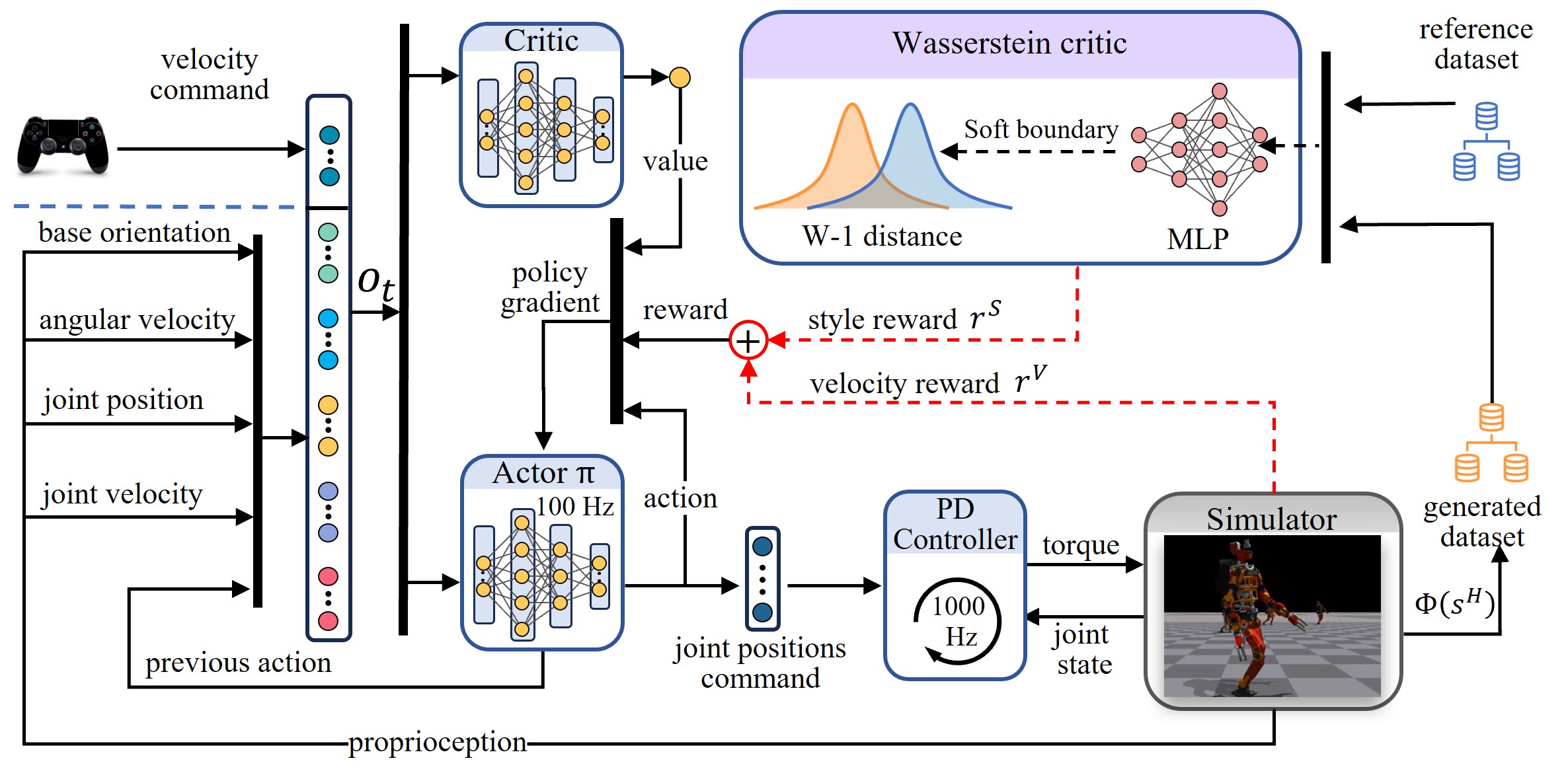} 
        \caption{\textbf{Wasserstein Adversarial Imitation Framework.} Given the robot's proprioceptive state and base velocity commands, the policy network predicts the joint position targets. A PD controller converts these targets into torques to actuate the robot. Using the reference motion dataset and policy-generated motion dataset, the Wasserstein critic updates its parameters through the soft-boundary Wasserstein-1 loss during training and predicts the style reward during roll-out. The style reward $r^S$ is combined with the velocity reward $r^V$ to guide policy training.}
        \label{fig_framework}
\end{figure*}

Our Wasserstein adversarial imitation learning framework, as illustrated in Figure \ref{fig_framework}, incorporates actor-critic networks, a Wasserstein critic, and a motor-level Proportional Derivative (PD) controller. The actor updates the network parameters using policy gradients derived from both environment rewards and the Wasserstein critic. The Wasserstein critic undergoes adversarial training based on the Wasserstein-1 distance complemented by a soft boundary loss. When given a user-defined velocity command, our setup enables humanoids to follow the velocity, ensuring smooth transitions in locomotion.
\subsection{Velocity-Conditioned Reinforcement Learning}
We formulate the humanoid locomotion control as a velocity-goal-conditioned\cite{chane2021goal} Markov decision process, with the velocity goal $v^{*}\sim p(v)\in\mathcal{V}$, state $s\in\mathcal{S}$, action $a\sim\pi(\cdot|s, v^*)\in\mathcal{A}$, reward $r = r(s,a,v^{*})$ and discount factor $\gamma \sim (0,1]$. The agent updates the decision policy $\pi$ through interactions with the surrounding environments to maximize the expected discounted return under the condition of the desired velocity,
\begin{equation}
        J(\pi)=\mathbb{E}_{v^* \sim p(v), \tau \sim p(\cdot \mid \pi ,v^*)}\left[\sum_t \gamma^t r\left(s_t, a_t, v^*\right)\right].
\end{equation}

The total reward terms are composed of two components:(1) velocity-tracking reward $r^V $, (2) style reward $r^S$,
\begin{equation}
        r_t=\mu_1 r^V + \mu_2 r^S,
\end{equation}
where $\mu_1$ and $\mu_2$ denote the combination weight on each term. Reward $r^V$ encourages the robot to follow the commanded CoM velocities, it is designed as the normalized exponential errors of linear velocity $v^*_{xy}$ and heading velocity $w^*_{z}$ separately,
\begin{equation}
        \begin{split}
        r^V = & \beta_1\exp \left(-\frac{\|v^*_{xy}-v_{xy}\|^2}{\lambda_l |v^*_{xy}|}\right) \\
              & + \beta_2\exp \left(-\frac{\|w^*_{z} - w_{z}\|^2}{\lambda_h |w^*_{z}|}\right),
        \end{split}
\end{equation}
$\beta_1$ and $\beta_2$ are hyper-parameters to control the importance of each tracking error. The parameters $\lambda_l$ and $\lambda_h$ are utilized to regulate the tracking precision. Smaller values of $\lambda_l$ and $\lambda_h$ encourage the humanoid robots to achieve better velocity-following precision but may make it more difficult for the policy to receive rewards at the beginning of training.

We model the actuation dynamics as a mass-damping system. A PD controller is employed to map the actions to desired torques with the target joint velocity always specified as 0.
% \begin{equation}
%         \tau=k_p(\hat{\theta}-\theta_t)-k_d(0-\dot{\theta_t}),
%         \end{equation}
% where $k_p$ and $k_d$ are proportional gain and derivative gain, $\hat{\theta}$ is the output target joint position from actor network, $\theta_t$ and $\dot{\theta_t}$ are joint position and velocity in current control step.

\subsection{Wasserstein Critic}
In adversarial imitation learning, it is pivotal for the discriminator to offer an appropriate distance metric between the generated data distribution $\mathcal{Q}$ and the reference data distribution $\mathcal{P}$. In vanilla GAIL\cite{ho2016generative}, the discriminator employs BCE loss, which has been shown to equate to minimizing the Jensen-Shannon Divergence\cite{ke2021imitation}. When there is no overlap between two high-dimensional data distributions, it can result in gradient vanishing, severely causing unstable training and mode collapse. IPMs have been proven as excellent distance measures on probabilities\cite{sriperumbudur2009integral},
\begin{equation}
        \Gamma_{\mathcal{F}}(\mathcal{P}, \mathcal{Q}):=\sup _{f \in \mathcal{F}}\left|\int_\mathcal{M} f d \mathcal{P}-\int_\mathcal{M} f d \mathcal{Q}\right|,  
\end{equation}
where $\mathcal{F}$ represents a class of real-valued bounded measurable functions on Manifold $\mathcal{M}$. When $\mathcal{F}=\left\{f:\|f\|_L \leq 1\right\}$, it forms the dual representation of Wasserstein-1 distance and the typical Wasserstein loss with gradient penalty \cite{gulrajani2017improved} becomes
\begin{equation}\label{equa:old_loss}
        \begin{split}
        \underset{\theta}{\arg\min}  &-\mathbb{E}_{x \sim P_r}\left[D_\theta(x)\right]+\mathbb{E}_{\tilde{x} \sim P_g}\left[D_\theta(\tilde{x})\right] \\
        &+ \lambda{\mathbb{E}}_{\hat{x} \sim P_{\hat{x}}}\left[\left(\left\|\nabla_{\hat{x}} D_\theta(\hat{x})\right\|_2-1\right)^2\right],
        \end{split}
\end{equation}
where $D_\theta(\cdot)$ denotes outputs from the Wasserstein critic. $x = \Phi\left(s^N\right)$ is the manually selected feature from $N$ consecutive motions $s^N$ that are sampled from the reference and generated motion distribution $P_r$ and $P_g$. $\hat{x}=\alpha x+(1-\alpha) \tilde{x}$ are samples obtained through random interpolation between the reference samples $x$ and generated samples $\tilde{x}$.

\textbf{Soft boundary constraint}. The Wasserstein critic network is used to approximate a cluster of Lipschitz-constrained functions with a linear combination architecture in the final layers. As a result, the output value is unbounded and unbiased \cite{zhang2020wasserstein}, \cite{kostrikov2018discriminator}. Utilizing original Wasserstein loss \eqref{equa:old_loss}, we observed drawbacks stemming from the unbounded values. At the early training stage, there are significant differences between generated samples and real data distributions, the critic's output for generated samples converges quickly to large negative values. This renders the style reward $r^{S}$ nearly zero, causing the policy to fail to learn natural motions. The unbounded value also introduces large standard deviations in style reward, which makes the training unstable. To limit the outputs from the Wasserstein critic, we modify the Wasserstein loss with a soft boundary constraint,
\begin{equation}\label{equa:wloss}
        \begin{split}
                \underset{\theta}{\arg\min}  &-\mathbb{E}_{x \sim P_r}\left[\tanh(\eta D_\theta(x))\right]\\ 
                &+\mathbb{E}_{\tilde{x} \sim P_g}\left[\tanh(\eta D_\theta(\tilde{x}))\right]\\
                & + \lambda \mathbb{E}_{\hat{x} \sim P_{\hat{x}}}\left[(\max \{0,\|\nabla_{\hat{x}} D_{\theta}(\hat{x})\|_2-1\})^2\right],
        \end{split}
\end{equation}
Where $\eta$ is a hyperparameter that controls the range of boundaries. Smaller $\eta$ means softer constraints that generate larger critic values. In practice, $\eta \sim (0.1, 0.5)$ is a proper range for selection. We apply a weaker gradient penalty \cite{petzka2018regularization} to further stabilize the training. Finally, the style reward is designed as $r^{S} = e^{ D_\theta(\tilde{x})}$.
% \begin{equation}
% As a result, the output value, which critics the distribution distance, is equivalent to the unbounded and unbiased \cite{zhang2020wasserstein}\cite{kostrikov2018discriminator} form,
% \begin{equation}
%         D_\theta(x) = \log (\sigma(x))-\log (1-\sigma(x)),
% \end{equation}
% where $\sigma(x)$ is the sigmoid function. 
\begin{figure*}[t]
        \centering
        \includegraphics[width=0.92\textwidth]{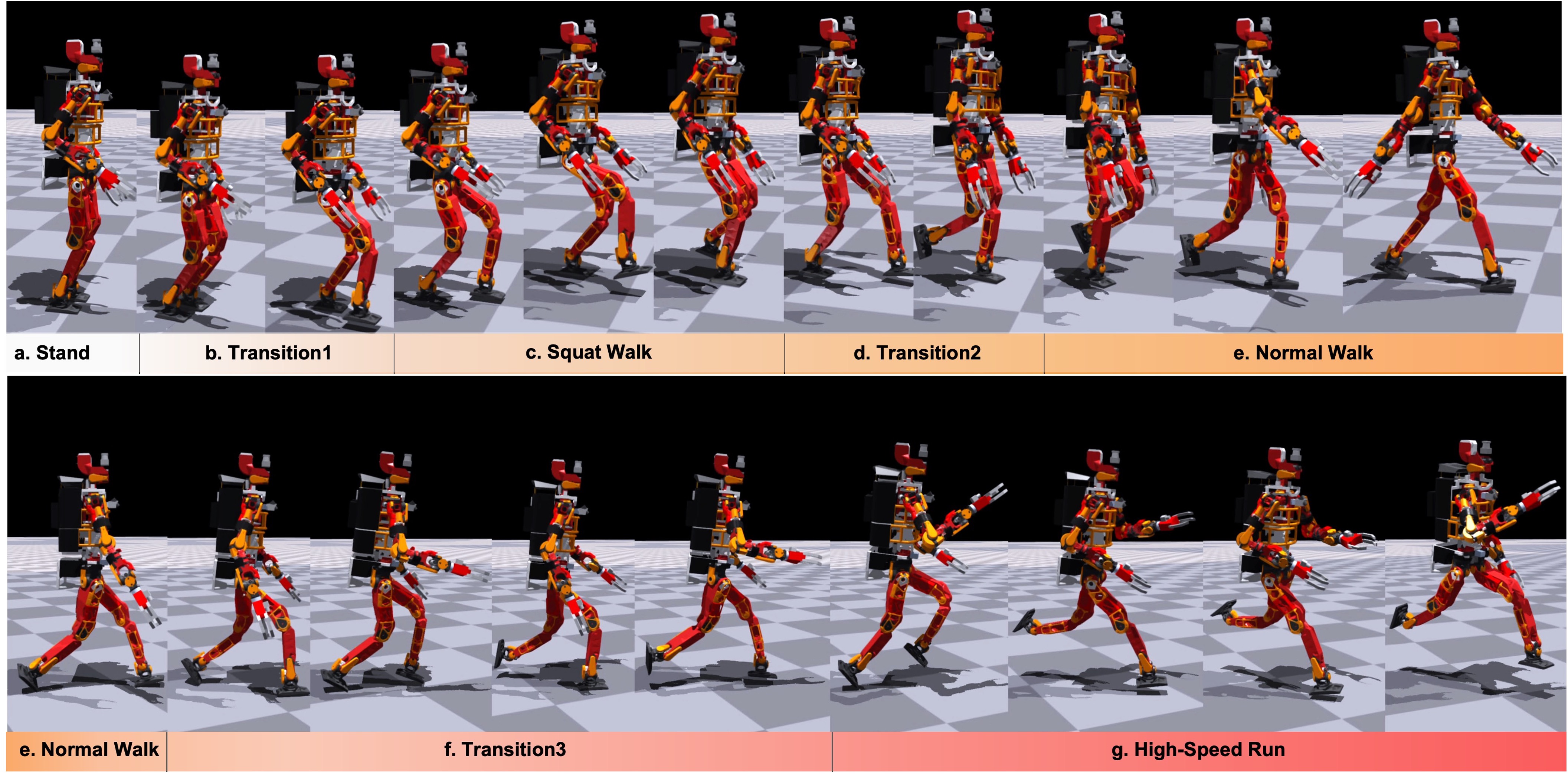} 
        \caption{Snapshots of various natural locomotion behaviors learned by the Humanoid JAXON. As the velocity command increases from 0 m/s to 5 m/s, the robot exhibits seamless transitions from standing to dynamic running. }
        \label{fig:natural_transition}
\end{figure*}

\section{Experiment}

\subsection{Implementation Details}

\textbf{Actor-Critic observation space}. The actor and critic networks share the same observation space. The observation space $O_{ac} \in \mathbb{R}^{102}$ consists of:
\begin{enumerate*}[label=(\roman*)]
        \item Base angular velocity $w_b \in \mathbb{R}^3$ expressed in base local frame.
        \item Velocity command $v^* \in \mathbb{R}^3$, including target linear velocity $v^*_{xy} \in [0, 5]$ m/s and heading velocity $w^*_z \in [-1,1]$ rad/s.
        \item The gravity vector $z_b \in \mathbb{R}^3$ expressed in base local frame.
        \item Current joint position $\theta \in \mathbb{R}^{31}$.
        \item Current joint velocity $\dot{\theta} \in \mathbb{R}^{31}$.
        \item Last-step actions $a_{t-1} \in \mathbb{R}^{31}$.
\end{enumerate*}

\textbf{Wasserstein-Critic observation space and action space}. The observation space $O_{wc}$ of Wasserstein critic consists of state-transition pairs $\Phi\left(s^N\right)= \left(s_i, \ldots, s_{i+N}\right) \in \mathbb{R}^{78\times \left(N+1\right)}$ in $N$ preceding time-steps. Each $s_i$ is represented in the same style feature space where the style features are carefully hand-selected. The motion style feature $s_i \in \mathbb{R}^{78}$ is composed of:
\begin{enumerate*}[label=(\roman*)]
    \item Base height $p_z \in \mathbb{R}$.
    \item Base linear velocity $v_b \in \mathbb{R}^3$ expressed in base local frame.
    \item Base angular velocity $w_b \in \mathbb{R}^3$ expressed in base local frame.
    \item The gravity vector $z_b \in \mathbb{R}^3$ expressed in base local frame.
    \item Joint position $\theta \in \mathbb{R}^{31}$.
    \item Joint velocity $\dot{\theta} \in \mathbb{R}^{31}$.
    \item Relative position of feet with base $r_{\text{feet}} \in \mathbb{R}^{6}$.
\end{enumerate*}
The corresponding action space $\mathcal{A} \in \mathbb{R}^{31}$ of policy is chosen as 31 target joint positions within the joint angle limitation.

\textbf{Reference motion dataset}. The reference motion dataset includes multiple locomotion patterns. Table \ref{tbl:training_dataset} depicts the Statistics details of the whole dataset used for training. Normal walk and squat walk are retargeted from CMU-MoCap dataset\cite{cmudata} and SFU-MoCap dataset\cite{sfudata}. The standstill motion is manually designed and the squat walk motion is recorded from the existing robot controller\cite{kojio2020footstep}. 

\textbf{Regularization terms and domain randomization}. To obtain a high-fidelity controller, we impose regularization penalties for large action jerks, significant joint torque, and acceleration. We also employ domain randomization on contact friction, restitution, joint friction, joint inertia, mass parameters, PD gains, and motor strength to avoid overfitting the environmental dynamics.

\textbf{Training details}. The actor, critic, and Wasserstein critic have the same MLP structures with [1024, 512, 256] hidden units and  ELU activation functions. Policies update via PPO\cite{schulman2017proximal} with a learning rate of $l = 3.0 e-5$ and around 30 hours of training in Isaac gym\cite{makoviychuk2021isaac} with an NVIDIA 3090Ti.
\setlength\extrarowheight{3pt}
\begin{table}[t]
\centering
\caption{Statistics of reference data}\label{tbl:training_dataset}
\begin{adjustbox}{width=\linewidth}
\begin{tabular}{cccc}
\hline
\textbf{Data Type} & \textbf{Duration(s)} & \textbf{Velocity(m/s)} & \textbf{Selection Probability} \\ \hline \hline
Stand              & 5.1                   & {[}0.0, 0.0{]}        & 0.15                        \\ \hline
Squat Walk     & 8.0                   & {[}0.2, 0.6{]}        & 0.20                         \\ \hline
Normal Walk        & 14.2                  & {[}0.9, 2.4{]}        & 0.35                        \\ \hline
Run                & 15.3                  & {[}2.9, 4.8{]}        & 0.30                         \\ \hline \hline
\textbf{Total}     & 42.6                  & {[}0.0, 4.8{]}        & 1.00                          \\ \hline
\end{tabular}
\end{adjustbox}
\end{table}
\subsection{Evaluation}
\begin{figure}[t]
        \centering
        \includegraphics[width=0.47\textwidth]{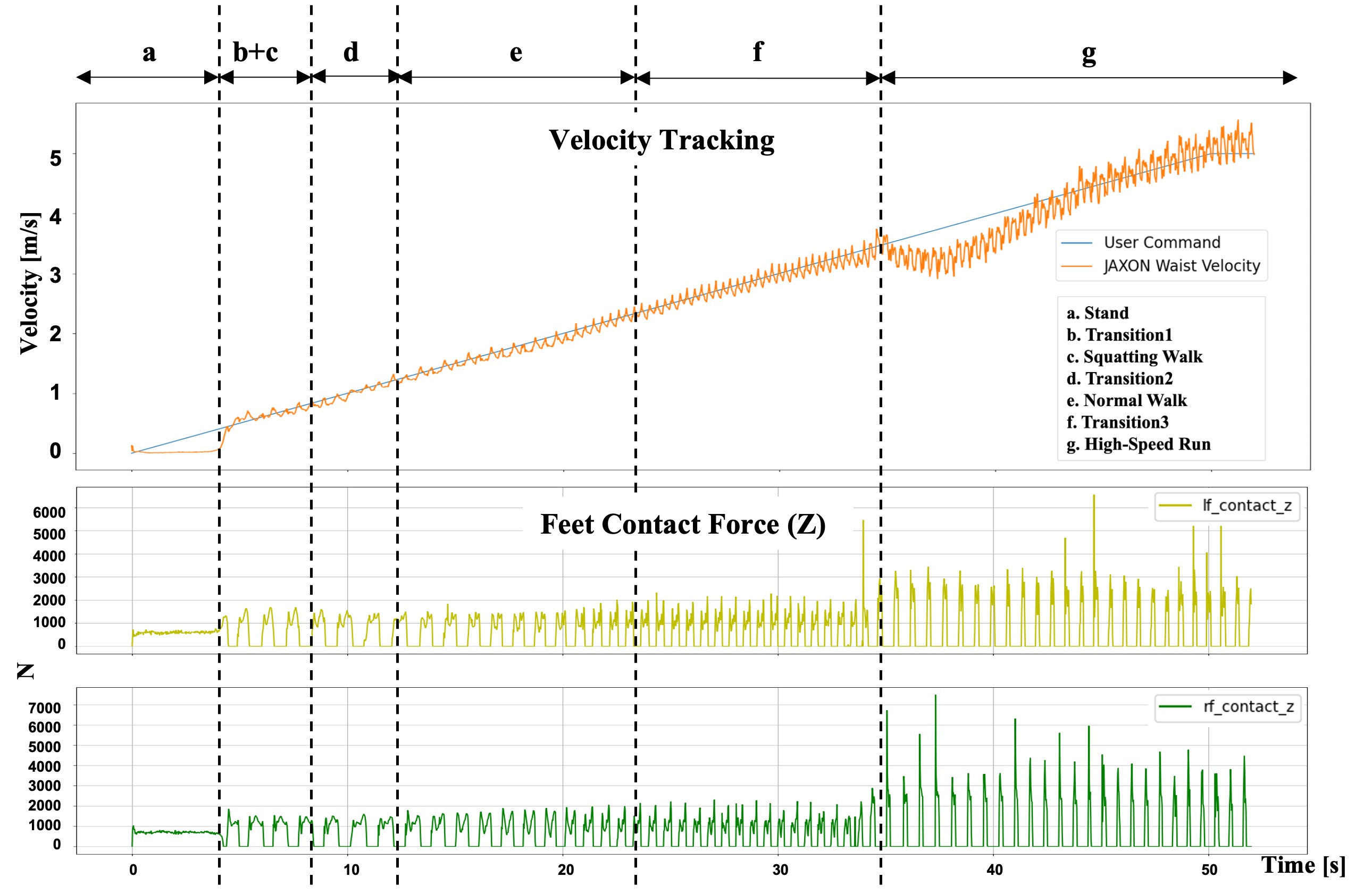} 
        \caption{Top: the velocity tracking curve, where the velocity command increases from 0.0 m/s to 5.0 m/s with a constant acceleration of 0.1 $\text{m/s}^2$. Middle and bottom: the contact forces in the z-direction for the left and right feet during the transition from standing to running.}
        \label{fig:vel_contact}
\end{figure}

\textbf{Natural locomotion and transition}. We examined the robot's ability to reproduce a range of natural locomotion behaviors from the reference dataset and to adapt to velocity commands. We set the initial desired velocity to 0 m/s and gradually increased it to 5 m/s with a constant acceleration of 0.1 $\text{m/s}^2$. Figure \ref{fig:natural_transition} presents a side view of JAXON robot's locomotion behaviors in response to changing velocities. The results indicate that our control policy not only captures diverse locomotion patterns from the reference dataset but also enables smooth transitions not present in the reference motions. The velocity-tracking curve and the z-direction feet contact force are shown in Figure \ref{fig:vel_contact}. The velocity tracking curve in Figure \ref{fig:vel_contact} demonstrates the robot's capability to closely follow the desired velocity, reaching speeds of up to 5 m/s. It's important to note that this velocity tracking refers to the average velocity over a gait cycle, as opposed to instantaneous velocity. As the speed increases, the variance in instantaneous velocity also increases. The contact force increases dramatically with the increase in speed. During the transition from walking stage $f$ to running stage $g$, there is a significant increase in stride frequency and a substantial decrease in contact time. As the robot transitions into the running gait pattern, we can observe the presence of the air phase.

\begin{figure}[t]
        \centering
        \includegraphics[width=0.49\textwidth]{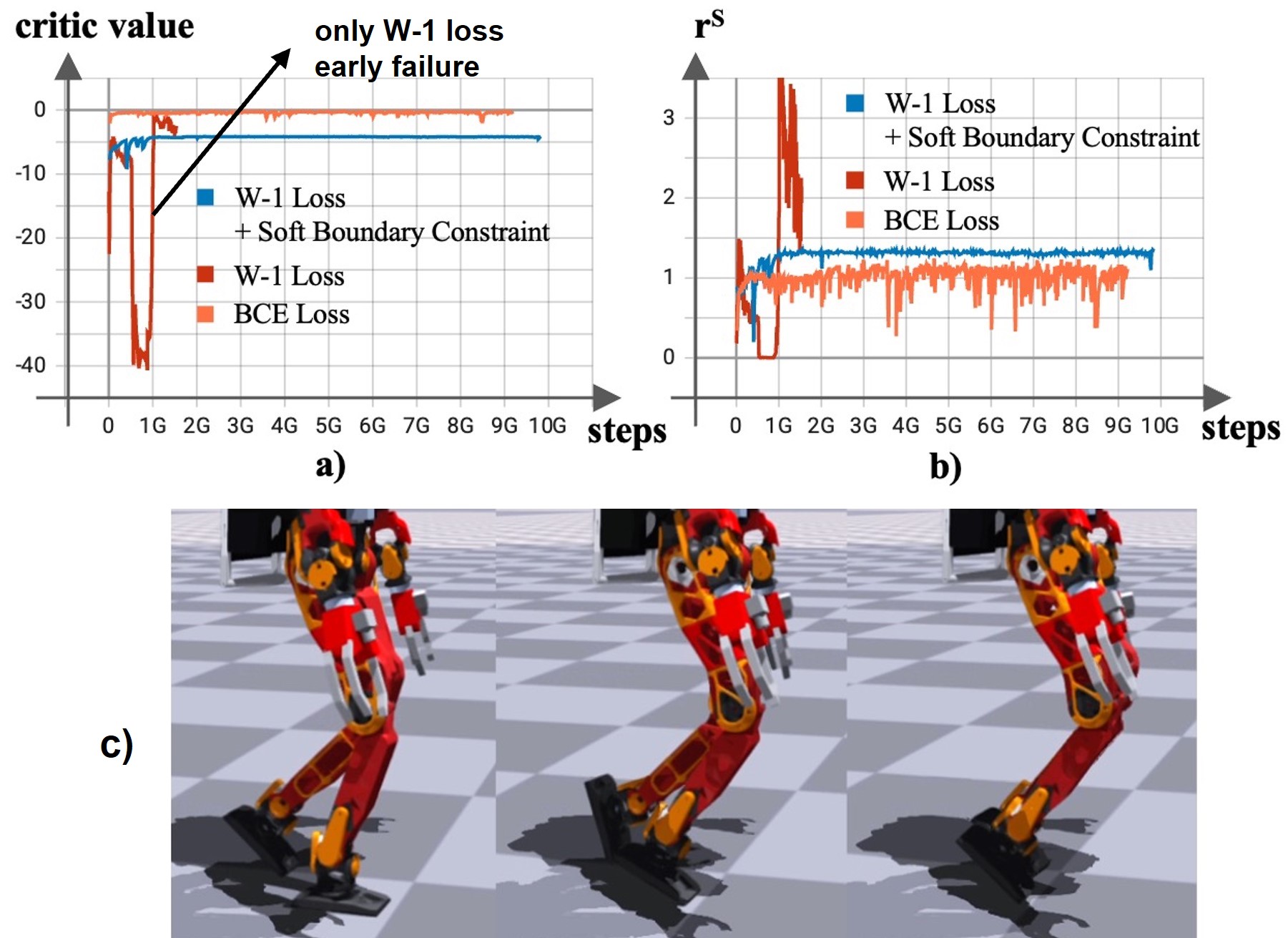} 
        \caption{a) Comparison of discriminator (critic) output values. b) Comparison of style reward values. Using only the Wasserstein-1 loss results in a wide range and significant fluctuations in both output and style reward, causing early-stage training failures. While employing the BCE loss keeps these values within a suitable range, it also leads to considerable relative fluctuations and susceptibility to mode collapse and unstable training. In contrast, the Wasserstein-1 loss with soft boundary constraint ensures both the output and style reward remain within an appropriate range and exhibit minimal fluctuations, leading to a more stable training process. c) An example of mode collapse with BCE loss is that the robot only learned a tiptoe walking gait close to the standing posture.}

        \label{fig:train_comparison}
\end{figure}

\textbf{Training stability and mode collapse}. To assess the utility of the
soft-boundary-constrained Wasserstein loss, we conducted three separate training sessions, each utilizing identical hyperparameters and random seeds, but varying discriminator loss types. As depicted in Figure \ref{fig:train_comparison}a and \ref{fig:train_comparison}b, the contrasts in discriminator (critic) outputs and style rewards are evident. With the W-1 loss, the critic’s output experiences significant fluctuations spanning a broad range. This causes rapid changes in the style reward \( r^S \) during the initial training phases, culminating in a failed training attempt. Conversely, while the discriminator output using the BCE loss remains between (0,1), it still exhibits considerable relative fluctuation, resulting in volatile changes to the style reward and destabilizing the training phase. Our novel soft-boundary-constrained Wasserstein loss effectively constrains the output value within a more acceptable range and also minimizes the fluctuation in style reward, thus enhancing training stability. Beyond stability, the Wasserstein critic delivers improved assessments of distributional distances, which ultimately curtails mode collapse and aberrant locomotion behaviors. 

\begin{figure}[t]
        \centering
        \includegraphics[width=0.49\textwidth]{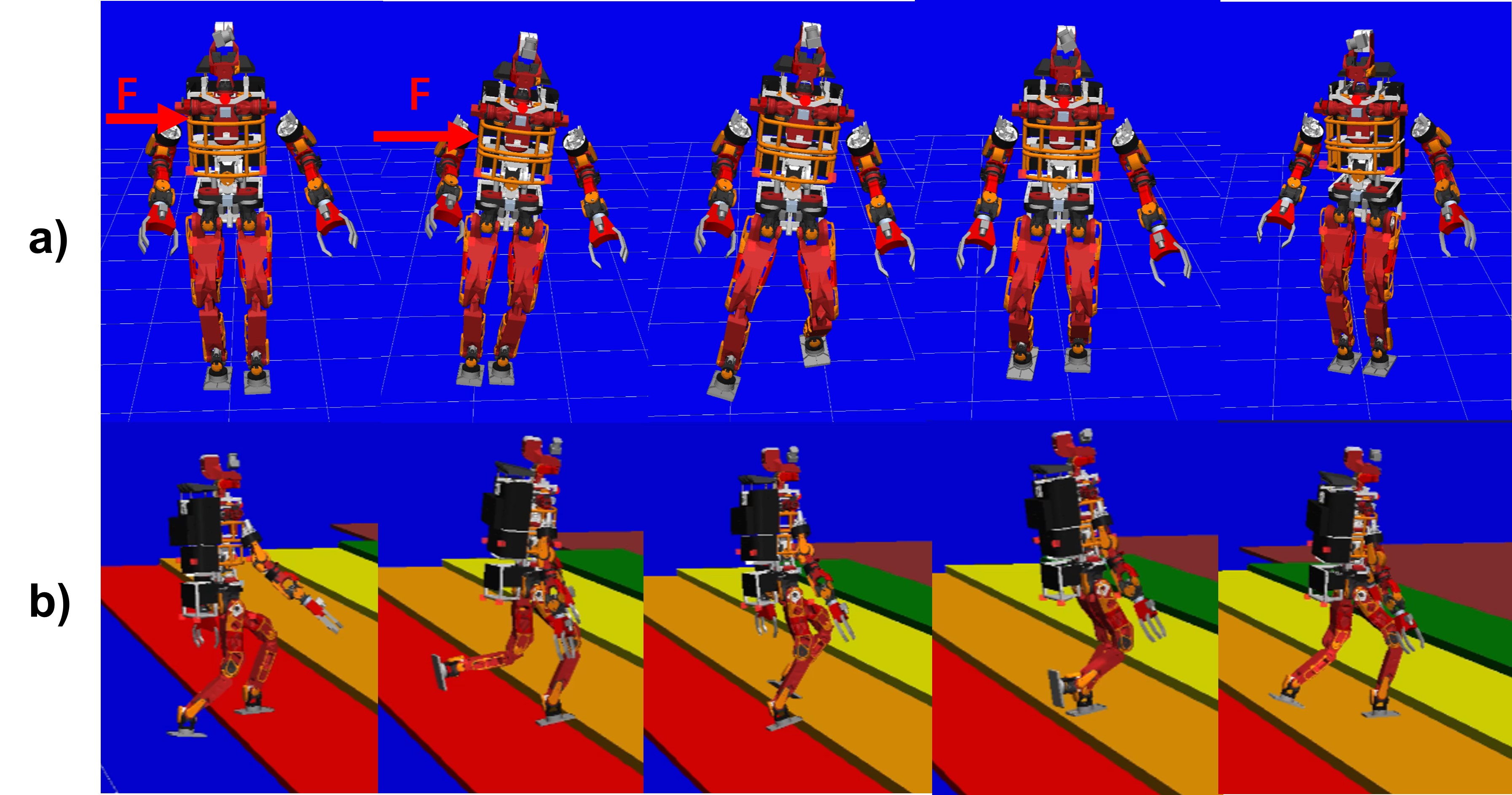} 
        \caption{Sim-to-sim robust test in high-fidelity Choreonoid simulator. a) Push-recovery: The robot takes one lateral step with its left foot to maintain balance. b) Stair-climbing: The robot navigates a set of stairs with each step height of 50mm.}
        \label{fig:go_stairs}
\end{figure}

\textbf{Sim-to-Sim robust test}. The Choreonoid\cite{nakaoka2012choreonoid}, integrated with real-time-control software Hrpsys, has been widely used in our previous work\cite{kakiuchi2015development}, \cite{kojio2019unified}, \cite{sato2022robust} and has proven to be a high-fidelity simulation environment with a small reality gap. We successfully transferred the policy from Isaac Gym to Choreonoid to facilitate future sim-to-real experiments. As depicted in Figure \ref{fig:go_stairs}, the controller demonstrates extraordinary robustness in push-recovery and blind stair-climbing tasks. 

\section{Conclusion}
In this work, we present a Wasserstein adversarial imitation learning system capable of acquiring a variety of natural locomotion skills from human demonstration datasets with diverse motion behaviors. We have detailed a unified primitive-skeleton motion retargeting method, proficient in efficiently mapping motions between skeletons with significant morphological differences. Our findings highlight the system's novel ability to seamlessly transition between unique locomotion patterns as the desired speed varies, even though such transition behaviors are conspicuously absent in the reference dataset. Further experiments validate that our proposed soft-boundary-constrained Wasserstein-1 loss significantly stabilizes the training process and reduces the risk of mode collapse. In the further, we aim to transfer this policy to real-world robots, to achieve versatile, natural, and dynamic locomotion for humanoids.

\bibliography{main}
\bibliographystyle{IEEEtran}

\end{document}